\documentclass{article}

\usepackage[final]{nips_2017}
\usepackage[utf8]{inputenc} % allow utf-8 input
\usepackage[T1]{fontenc}    % use 8-bit T1 fonts
\usepackage{hyperref}       % hyperlinks
\usepackage{url}            % simple URL typesetting
\usepackage{booktabs}       % professional-quality tables
\usepackage{amsfonts}       % blackboard math symbols
\usepackage{nicefrac}       % compact symbols for 1/2, etc.
\usepackage{microtype}      % microtypography

\usepackage{graphicx}
\usepackage{amsmath}       % eqref

\title{Household poverty classification in data-scarce environments: a machine learning approach}

\author{
  Varun Kshirsagar \\
  Innovations for Poverty Action\\
  Washington, DC 20005 \\
  \texttt{vkshirsagar@poverty-action.org} \\
  \And
  Jerzy Wieczorek \\
  Department of Statistics \& Data Science\\
  Carnegie Mellon University\\
  Pittsburgh, PA 15213 \\
  \texttt{jerzy@stat.cmu.edu} \\
  \And
  Sharada Ramanathan \\
  Innovations for Poverty Action\\
  Singapore \\
  \texttt{sramanathan@poverty-action.org} \\
  \And
  Rachel Wells \\
  Innovations for Poverty Action\\
  Washington, DC 20005 \\
  \texttt{rwells@poverty-action.org} \\
}

\begin{document}

\maketitle

\begin{abstract}
We describe a method to identify poor households in data-scarce countries by leveraging information contained in nationally representative household surveys. It employs standard statistical learning techniques---cross-validation and parameter regularization---which together reduce the extent to which the model is over-fitted to match the idiosyncracies of observed survey data. The automated framework satisfies three important constraints of this development setting: i) The prediction model uses at most ten questions, which limits the costs of data collection; ii) No computation beyond simple arithmetic is needed to calculate the probability that a given household is poor, immediately after data on the ten indicators is collected; and iii) One specification of the model (i.e.\ one scorecard) is used to predict poverty throughout a country that may be characterized by significant sub-national differences. Using survey data from Zambia, the model's out-of-sample predictions distinguish poor households from non-poor households using information contained in ten questions.
\end{abstract}

\section{Introduction}
\label{sec:intro}
Any organization whose mission is to serve the poor must have a way to identify poor households as potential clients or customers, as well as to track poverty rates in the communities served. However, in order to decisively identify households that are poor, the requisite assessment of household characteristics and assets can be expensive and burdensome, requiring several hours of interviewer time per household.

A faster alternative, which maintains rigor while placing less burden on the respondent and requiring less interviewer time, is to use a \emph{proxy-means test} (PMT). A PMT is typically a scorecard or formula which uses only a small number of questions, perhaps between 10 and 30, to estimate the probability that a household is poor. Such a PMT can be developed using data from a nationally-representative household survey: select a subset of variables from the survey, estimate a model using these variables to predict poverty at the household level, translate the fitted model into a scorecard, and validate the resulting PMT using held-out data and field tests (e.g.\ Schreiner 2007, Schreiner 2010). For instance, the PMT for a particular country, year, and poverty line may ask simple questions such as ``What material is your roof made out of?'' or ``Does every child between 6 and 12 attend school?''.

Our novel approach to constructing a PMT uses modern machine learning methods for variable selection and regression, as described in Section~\ref{sec:methodology}.
Traditional econometric approaches to constructing PMTs typically use simple stepwise logistic regression (Basurto et al.\ 2016; Brown et al.\ 2016). Such classical greedy algorithms are known to have major shortcomings such as high variability in the model-selection step and in the consequent prediction errors. This concern is most acute when predictors are highly correlated, as they tend to be in the national household expenditure surveys used to construct PMTs.
In contrast, we believe our modern methodology can choose more stable variable subsets and estimate scorecards with better out-of-sample accuracy than traditional approaches.

\subsection{Poverty Probability Index (PPI)}

In this section, we describe a new methodology for constructing a PMT known as the Poverty Probability Index (PPI). PPI scorecards are currently available for 60 countries which together include over 90\% of the world's poor living below 1.90 per person per day PPP \$. The product is used by nearly 600 organizations around the world---NGOs, corporations, social enterprises---to estimate the poverty level of the households they serve.
This poverty measurement tool is developed by Innovations for Poverty Action under the support and guidance of the PPI Alliance. In designing the new approach to product development reported here, we have had to satisfy several constraints based on historical user expectations and desired objectives. We have retained a very similar front-end (Schreiner 2007, Schreiner 2010), so that users did not need to change their approach. However, in contrast to the previous methodology---which employs stepwise regression and human judgment for variable selection, a logistic regression to estimate scores, and a non-parametric approach to assign probabilities to scores---we employ more conventional cross-validation and parameter regularization techniques to inform both model selection and estimation. Further, also in contrast to the previous methodology, as discussed below, our approach is designed to provide estimates that are accurate across different segments of a diverse country.\footnote{This was a major criticism of the previous methodology (e.g.\ Diamond et al.\ 2016).}

\paragraph{User expectations} Past PPI scorecards have incorporated 10 questions or variables (sometimes ranging between 8 and 12 variables), typically taking around ten to fifteen minutes to complete, instead of the several hours required for an household interview that provides the poverty measures we estimate. The 10 questions used by the PPI vary by country, but those same 10 questions are used within all sub-national regions and at all poverty lines. This way, interviewers can be trained to apply the PPI across a country. For this reason, we continue producing PPIs that use one set of 10 variables per country. We will also provide estimates of the PPI's accuracy when applied separately to the sub-national areas for which inference is supported by the survey design.

Further, not all organizations that use the PPI have access to digital survey instruments. To enable simple pencil-and-paper use, the original PPI scorecards were designed as additive models with categorical predictors. Every possible response to each of the 10 questions is assigned a weight (or scorecard points) of zero or a positive integer. Field interviewers collect a household's responses, sum up the weights for each selected response to compute a household score, and check this score against a printed lookup table with the estimated probability of poverty associated with every score. Again, for the sake of consistency, our updated PPI continues to rely on this approach. This constrains us to use variants of logistic regression (without interactions, so that interviewers never need to multiply weights).\footnote{We are not constrained by computational ability to \emph{fit} a model to training data, only to \emph{deploy} it in the field.}

We begin with a candidate set of questions that, depending on the country and survey, vary in number from 30 to 100---so we still require a principled variable-selection algorithm (selecting 10 questions from 30 involves more than 2 million combinations, while selecting 10 questions from 50 involves more than 10 billion combinations of 10 questions). Finally, judgment (incorporated through field tests or desk evaluations from enumerators with local country experience) is needed to confirm that the selected questions can be answered in a reasonable time and can be administered by interviewers, who receive a day or two of training, but who often lack specialized training in administering survey questions.

\paragraph{Desired objectives}
The estimated model should be able to make accurate out-of-sample predictions. In particular, the predictions for poor and non-poor households should be well-separated. When we set a threshold on the predicted probability (classifying households below the threshold as non-poor, and above the threshold as poor), we wish to minimize both inclusion and exclusion errors: we should not include a non-poor household in our targeting, nor should we exclude a poor household.

In addition, predictive accuracy must be maintained across different segments of a diverse country. Both variable-selection and model-fitting must be stable across subsets of the data, since the same PPI product will be used across sub-national regions. This objective drives the design of our statistical learning approach and the decision to evaluate model performance by regions, not just nationally.

Each country's PMTs should be updated over time as conditions change and new household-survey data becomes available. Our approach is designed to be as fully automated as possible, to allow low cost and rapid development of the product across time, countries, and poverty lines.
\subsection{Data sources and survey design}

Data used to fit and evaluate a PPI scorecard come from national household expenditure or income surveys, using the most recent survey available. Such surveys are often collected by a national statistical agency in order to monitor demographic and socio-economic characteristics of the population.
In this paper, we use data from the 2015 Zambia Living Conditions Measuring Survey (LCMS), which was carried out by the Zambian Central Statistical Office.

Such national surveys generally use a complex survey design, often involving two-stage stratified sampling.
The released datasets usually include inverse-probability survey weights $w_i$ for each observation $i$. Hence, although our methodology relies on simple random sampling from the observed dataset to form bootstrap resamples and cross-validation folds, our elastic net regression coefficient estimates do make use of the survey weights.

% Such use of inverse-probability weights is recommended by McConville et al. (2017) for linear regression with the lasso. We show that the use of survey weights for logistic regression, as in Section 8.4 of Heeringa et al. (2010), is equivalently appropriate with the elastic net penalty.

Specifically, we fit weighted elastic net logistic regressions using the R package \texttt{glmnet} (Friedman et al.\ 2010), which solves the following penalized pseudo-loglikelihood problem:
\begin{equation}
\label{e-net}
\min_{\beta_0,\beta} - \frac{1}{n} \sum_{i =1}^n \Big[w_i \cdot l(\beta_0,\beta; y_i, x_i) \Big] + \lambda\Big[(1 - \alpha)\|\beta\|^{2}_{2} + \alpha\|\beta\|_{1}\Big]
\end{equation}
For logistic regression, observation $i$'s contribution to the loglikelihood is
\begin{equation}
\label{logistic-loglik}
l(\beta_0,\beta; y_i, x_i) =  y_i[\beta_0 + \beta^{T}x_{i}] - \log[1 + e^{\beta_0 +\beta^{T}x_{i}}]
\end{equation}
With $\lambda=0$, the unpenalized version of \eqref{e-net} reduces to the standard pseudo-loglikelihood for survey-weighted logistic regression, using the pseudo-maximum likelihood estimation approach (PMLE) described for instance in Section 8.4 of Heeringa et al.\ (2010).
The prefix ``pseudo'' indicates that we are maximizing a survey-weighted estimate of the likelihood, not the usual likelihood function.

For penalized regression, McConville et al.\ (2017) have studied survey-weighted linear regression with a lasso penalty. They recommend solving the same optimization problem as in \eqref{e-net}, but they only consider $\alpha=1$ and the linear regression loglikelihood $l(\beta_0,\beta; y_i, x_i) = \left[y_i-(\beta_0 +\beta^{T}x_{i})\right]^2$.
Survey-weighted logistic regression with an elastic net penalty does not appear to have been explicitly studied, but our use of \eqref{e-net} with \eqref{logistic-loglik} is the natural generalization of what McConville et al.\ recommend for the linear lasso case.

\section{Methodology}
\label{sec:methodology}
Our proposed method consists of several steps: variable selection, fitting the selected model, and translating the model into scorecard format. Our variable-selection is designed for lower variability than traditional stepwise methods, and penalized coefficient estimates are meant to prevent overfitting.

\paragraph{Bootstrap variable selection}
In many countries, due to sub-national regional differences in poverty characteristics, variable selection using a nationally representative sample may lead to model specifications that perform poorly within separate sub-national regions. We address this concern using an approach adapted from Section 6.2 of Hastie et al. (2015), similar in spirit to the stability selection algorithm of Meinshausen and Buhlmann (2010). Within each of many $m$-out-of-$n$ bootstrap samples, we use survey-weighted elastic net logistic regression (cross-validated over the overall penalty strength $\lambda$) to select an active variable set by solving \eqref{e-net}.
% \begin{equation}
% \label{e-net}
% \min_{\beta_0,\beta} - \frac{1}{N} \sum_{i =1}^N \Big[ y_i[\beta_0 + \beta^{T}x_{i}] - log[1 + e^{\beta_0 +\beta^{T}x_{i}}] \Big] + \lambda\Big[(1 - \alpha)\|\beta\|^{2}_{2} + \alpha\|\beta\|_{1}\Big]
% \end{equation}

Afterwards, we choose $\hat{q}$, the set of 10 questions selected most frequently (i.e.\ having the lowest occurrence of zero weights across the bootstrap samples). Since our 10 questions are associated with poverty over many different sub-samples, they should allow us to fit a model that works well both nationally and within sub-national segments of a diverse country.

In addition to our variable selection procedure discussed above, we allow for sub-national differences in poverty by including sub-national region indicator (dummy) variables. We enforce that the model always includes these variables by assigning no elastic-net penalty to their coefficients, equivalent to having a separate un-penalized intercept term for each region.\footnote{For each observation $i$, let $r_i$ be the vector of region dummies and $z_i$ be the vector of all other predictor variables, so that $x_i=\{r_i,z_i\}$ and $\beta = \{\beta_r, \beta_z\}$. Then we solve a variant of \eqref{e-net} where the penalty is only applied to $\beta_z$, not all of $\beta$.}

This entire variable selection step is nested inside an outer cross-validation loop for the $l_1$-vs-$l_2$ penalty mixing parameter $\alpha$, along with the next step of model estimation.

\paragraph{Elastic net estimation}
Using the 10 variables chosen in the first step, we fit a new survey-weighted elastic net logistic regression to all observations in the dataset. Again, $\alpha$ is selected by an outer cross-validation loop over this and the previous step, but we use internal cross-validation to choose $\lambda$.
The resulting estimate is a regularized logistic model, additive in the 10 selected predictors.

\paragraph{Estimating the probability of poverty}
The estimated probability that a household is poor follows for the logistic regression that uses just 10 questions as predictors. 
\begin{equation}
\hat{P}(y_i =1 | x_i=\{r_i, z_i\}) = \Big[1 + e^{-\left[\hat{\beta}_0 + \hat{\beta}_{r}^T r_i + \hat{\beta}_z^Tz_i^{\hat q}\right]}\Big]^{-1}
\end{equation}
One may interpret the information contained in the ten questions as engendering a deviation from the expected regional poverty rate. If no household-level information is available, then the best estimate of the probability that a household is poor is the prevailing poverty rate in the region. However, when information contained in ten questions is available, this estimated probability can range from 0 to 1.

\paragraph{Translating the fitted model into a scorecard and lookup table}
With the data for household $i$, we can compute a response on the scale of the linear predictors: $\hat{\beta}^Tx_i^{\hat q}$. This response could be fed into a logistic function to compute the predicted probability of poverty, but not with pencil-and-paper arithmetic.

Instead, we shift, scale, and round the estimated logistic regression coefficients $\hat{\beta}$ into weights $\tilde{\beta}$, such that all valid combinations of the linear predictors correspond to integer scores between 0 and 100. First, each variable's baseline level is chosen so that high scores indicate a low probability of poverty. In other words, a poor household is defined as having characteristics consistent with a base response to every question. Then $S^{max}$ is the maximum score of the unadjusted coefficients of the selected questions ($x^{\hat q}_i$), while $S_i$ is the adjusted score that ranges from 0 to 100.
\begin{equation}
S^{max} = \max_i \hat{\beta}^{T} x^{\hat q}_i
\end{equation}
\begin{equation}
\tilde{\beta} = \mathrm{round}\Big[\frac{100}{S^{max}}\cdot \hat{\beta}\Big]
\end{equation}
The adjusted score is always between 0 and 100 points:
\begin{equation}
S_i =  \tilde{\beta}^{T} x^{\hat q}_i
\end{equation}
We tabulate the corresponding logistic function output for each possible integer score. This provides a lookup table with a score between 0 and 100 and a corresponding probability of poverty.

For instance, our estimated scorecard for Zambia includes questions such as ``What material was used to construct the roof?'' The weights associated with possible answers are 0 for Thatched, 7 for Iron Sheets, and 10 for Cement/Asbestos/Other, showing that a Thatched roof tends to be associated with poverty while a Cement roof does not.
Field interviewers find the score $S_i$ simply by summing up such integer weights for each question's selected response, then look up the score in a table, never computing a logistic function directly.

In this way, we are able to fit a computationally intensive modeling process on centralized data, yet deploy pencil-and-paper prediction models for poverty targeting in the field.

\section{Results}
\label{sec:results}
Using a random 2:1 train:test split, we train the modeling process above and then evaluate our predictions on the held-out data. We evaluate prediction quality nationally, across sub-national regions, and across consumption deciles.

\begin{figure}[h]
  \centering
  \includegraphics[width=70mm, height=70mm]{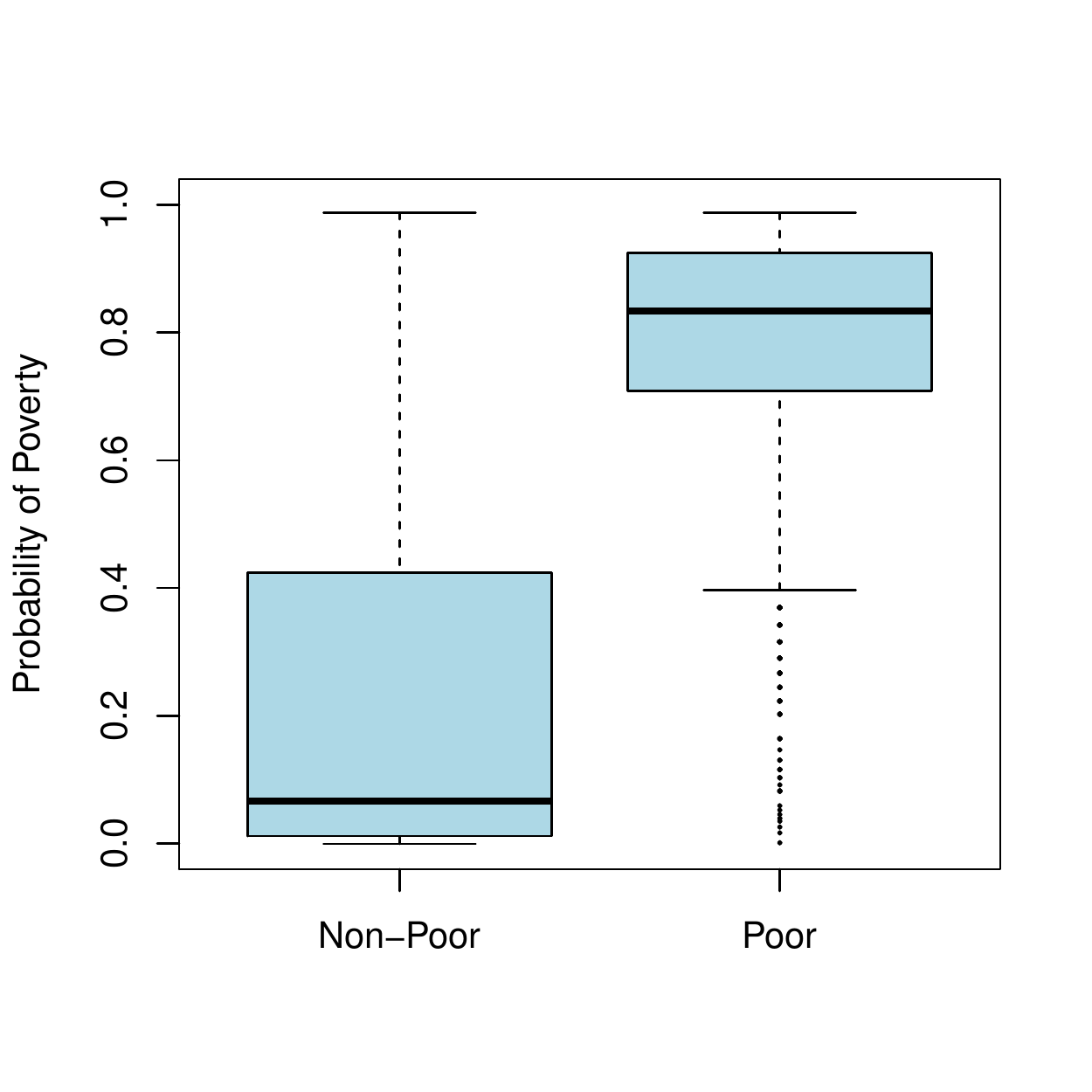}
  \caption{Out-of-sample predicted probability of poverty: Zambian national poverty line}
  \label{fig:box basic}
\end{figure}

It is worth beginning with the full national sample. Figure~\ref{fig:box basic} shows box plots of out-of-sample predictions for non-poor and poor households. The 75th percentile of the predicted probability of poverty for a poor household is around 0.4, substantially lower than the 25th percentile of the predicted probability for a poor household (around 0.75). A rule that assigns program benefits to households with a predicted probability of greater than 0.6 will exclude more than three quarters of non-poor households and exclude less than a quarter of poor households (include more than three quarters of the poor households).

\begin{figure}[h]
  \centering
  \includegraphics[width=128mm, height=80mm]{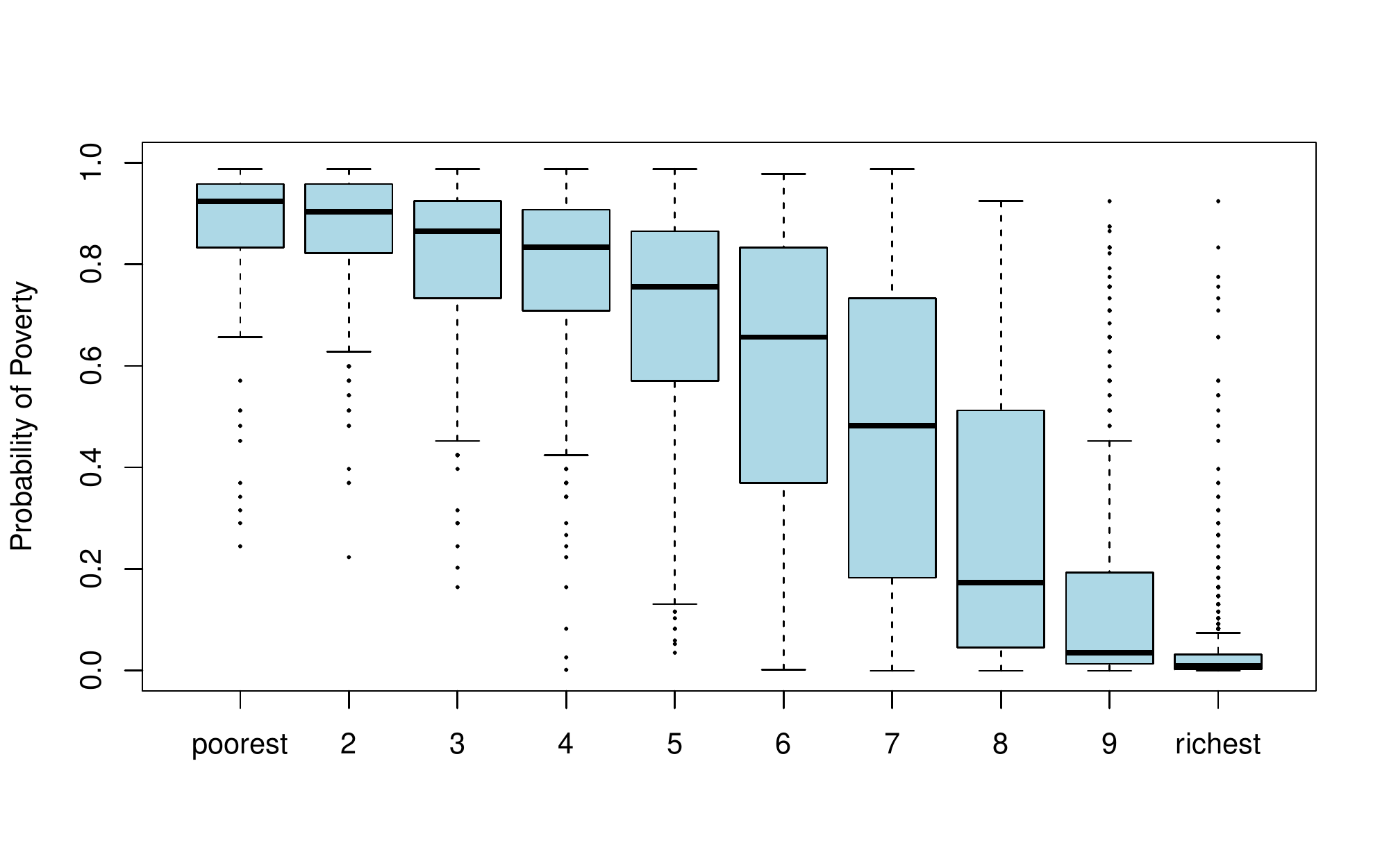}
  \caption{Out-of-sample predicted probability of poverty across consumption deciles}
  \label{fig:box deciles}
\end{figure}

Further, Figure~\ref{fig:box deciles} shows box plots of out-of-sample predictions by consumption decile. Household-level consumption---an input to our definition of poverty status---is derived from the cost of consuming a basket of goods and services consistent with a given standard of living, adjusted for household size and for local prices.
As expected, Figure~\ref{fig:box deciles} shows that households in the 1st decile (lowest 10\%) of the consumption distribution have the highest predicted probability of poverty, while households in the highest 10th decile are generally predicted to be non-poor. Further we see that there are substantial differences in predicted probability even between deciles at the middle of the distribution (e.g. between the fifth and sixth deciles).

\begin{figure}[h]
  \centering
  \includegraphics[width=128mm, height=80mm]{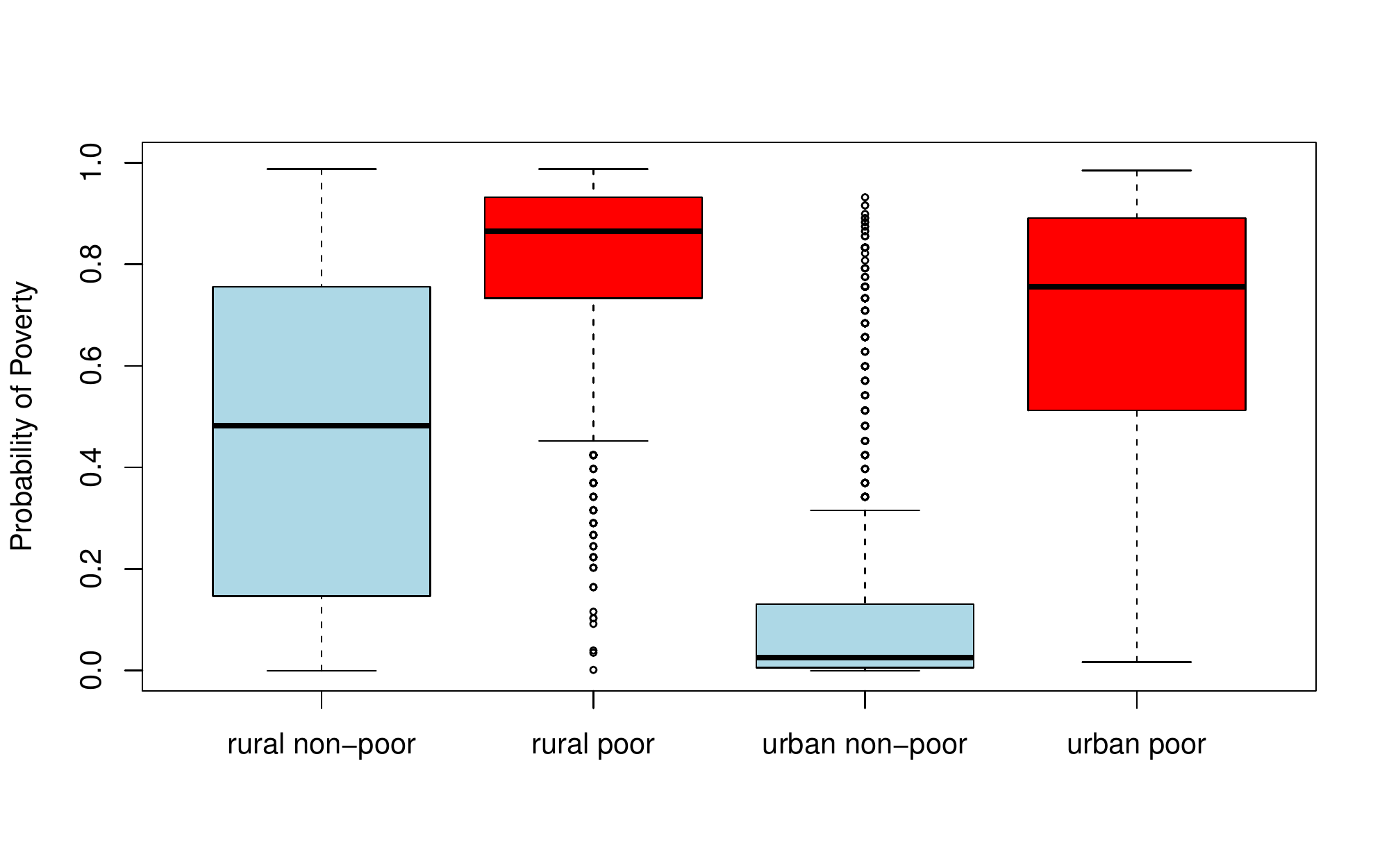}
  \caption{Out-of-sample predicted probability of poverty across urban-rural location and poverty status}
  \label{fig:box urban}
\end{figure}

Given that urban and rural livelihoods are different, it is challenging to select 10 questions that accurately predict poverty across both segments of a country. Figure~\ref{fig:box urban} shows that our methodology is able to successfully delineate poor households from non-poor in both rural and urban areas.

\begin{figure}[h]
  \centering
  \includegraphics[width=128mm, height=80mm]{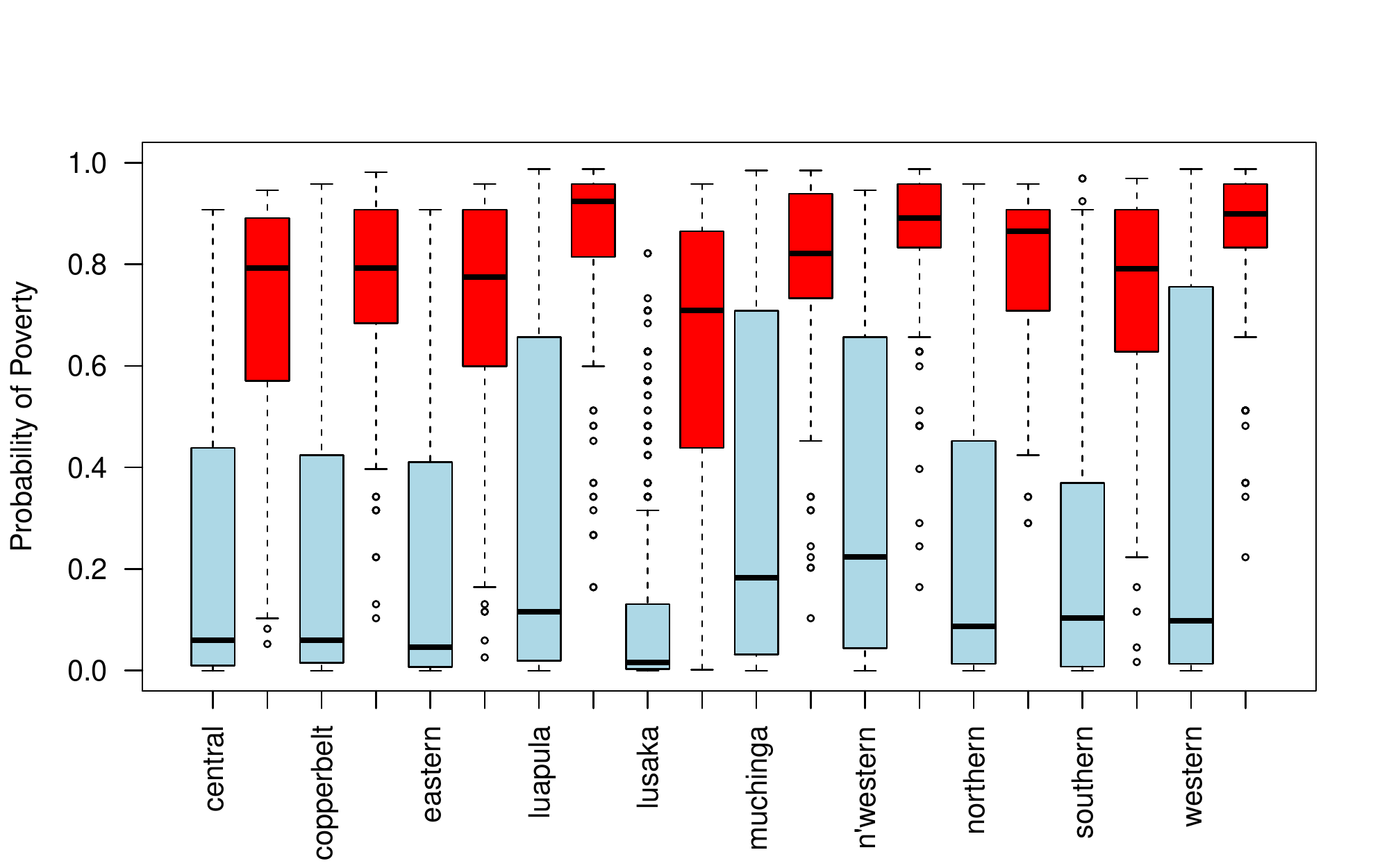}
  \caption{Out-of-sample predicted probability of poverty across provinces and poverty status}
  \label{fig:box province}
\end{figure}

Finally, we evaluate the predicted probability of poverty among poor and non-poor households, by sub-national region.  In Zambia, the main administrative regions are called provinces. There are significant economic differences between these provinces. For example, the Southern province is mainly a remote food surplus area, while Lusaka is mainly urban, and the Copperbelt has an economy that depends on the mining industry. Figure~\ref{fig:box province} shows that our methodology is able to separate (out-of-sample) poor households from non-poor households across all ten provinces in Zambia. However, there are some differences. Lusaka and the Southern province exhibit the greatest separation, while Muchinga is less accurate.

Overall, our methodology leads to reasonable separation between poor and non-poor households nationally and across several breakdowns (by
consumption deciles, sub-national regions, and urban-rural locations). The results are also very close to those from a full logistic regression model with no variable selection, showing that limiting the additive model to only 10 variables does not impair performance.

\section{Future work}
\label{sec:future}
Could models with even more capacity perform substantially better? We will compare our results to approaches that involve interaction between predictors, such as random forests, to diagnose whether, and how much, the additivity constraint impairs our scorecards.
We will also explore whether performance could be improved by fitting different scorecards across sub-national regions and/or poverty lines, borrowing strength by using hierarchical models and/or multinomial regression, respectively.
Finally, we will revisit open questions about machine learning with survey data: How should the bootstrap or cross-validation subsamples account for the survey design?

\subsubsection*{Acknowledgments}
We thank Julie Peachey (the PPI Director) and the PPI Alliance for useful comments and suggestions, as well as institutional support. We have also received valuable feedback from participants at presentations at Innovations for Poverty Action and the World Bank.
% Use unnumbered third level headings for the acknowledgments. All acknowledgments go at the end of the paper. Do not include acknowledgments in the anonymized submission, only in the final paper.

\section*{References}

\medskip

\small

Basurto, P., Dupas, P.\ \& Robinson, J.\ (2016) ``Decentralization and efficiency of subsidy targeting: evidence from chiefs in rural Malawi.'' {\it NBER Working Paper No.\ 23383}.

Brown, C., Ravallion, M.\ \& Van De Walle, D.\ (2016) ``A poor means
test? Econometric targeting in Africa.'' Technical Report, National Bureau of Economic Research.

Diamond, A., Gill, M.,  Dellepiane, M.A.R., Skoufias, E., Vinha, K.\ \& Xu, Y.\ (2016) ``Estimating poverty rates in target populations: An assessment of the simple poverty scorecard and alternative approaches.'' {\it World Bank Policy Research Working Paper 7793}.

Friedman, J., Hastie, T., \& Tibshirani, R.\ (2010). ``Regularization Paths for Generalized Linear Models via Coordinate Descent.'' {\it J.\ of Statistical Software} {\bf 33}(1):1-22.

Hastie, T., Tibshirani, R.\ \& Wainwright, M.\ (2015) {\it Statistical Learning with Sparsity: The Lasso and Generalizations.} Chapman \& Hall/CRC.

Heeringa, S.G., West, B.T.\ \& Berglund, P.A.\ (2010) {\it Applied Survey Data Analysis.} Chapman \& Hall/CRC.

McConville, K.S., Breidt, F.J., Lee, T.C.M.\ \& Moisen, G.G.\ (2017) ``Model-assisted survey regression estimation with the lasso.'' {\it J.\ of Survey Statistics and Methodology} {\bf 5}(2):131-158.

Meinshausen, N.\ \& Buhlmann, P.\ (2010) ``Stability selection.'' {\it JRSS(B)} {\bf 72}(4):417-473.

Schreiner, M.\ (2007) ``A simple poverty scorecard for the Philippines.''   {\it Philippine J.\ of Development} {\bf 34}(2):43-70.

Schreiner, M.\ (2010) ``A simple poverty scorecard for Pakistan.''   {\it J.\ of Asian and African Studies} {\bf 45}(3):326-349.

\end{document}